\documentclass[10pt,twocolumn,letterpaper]{article}

\usepackage{cvpr}

\usepackage{algorithmic}
\usepackage{amsmath}
\usepackage{algorithm}
\usepackage{multirow}
\usepackage{pifont}
\usepackage[T1]{fontenc}

\usepackage{array}
\usepackage[percent]{overpic}

\definecolor{cvprblue}{rgb}{0.21,0.49,0.74}
\usepackage[pagebackref,breaklinks,colorlinks,allcolors=cvprblue]{hyperref}

\def\paperID{19123} 
\def\confName{CVPR}
\def\confYear{2026}

\title{BrepGaussian: CAD reconstruction from Multi-View Images with Gaussian Splatting}

\author{
Jiaxing Yu$^{1}$, Dongyang Ren$^{1}$, Hangyu Xu$^{1}$, Zhouyuxiao Yang$^{1}$,\\ Yuanqi Li$^{1}$\footnotemark[2] , Jie Guo$^{1}$, Zhengkang Zhou$^{2}$, Yanwen Guo$^{1}$\\
{\small $^{1}$State Key Laboratory of Novel Software Technology, Nanjing University}\\
{\small $^{2}$Nanjing Urban Construction Tunnel\& Bridge Intelligent Management Co., Ltd.}
}

\begin{document}

\twocolumn[{%
  \renewcommand\twocolumn[1][]{#1}%
  \maketitle

  \begin{center}

    \begin{overpic}[width=\textwidth]{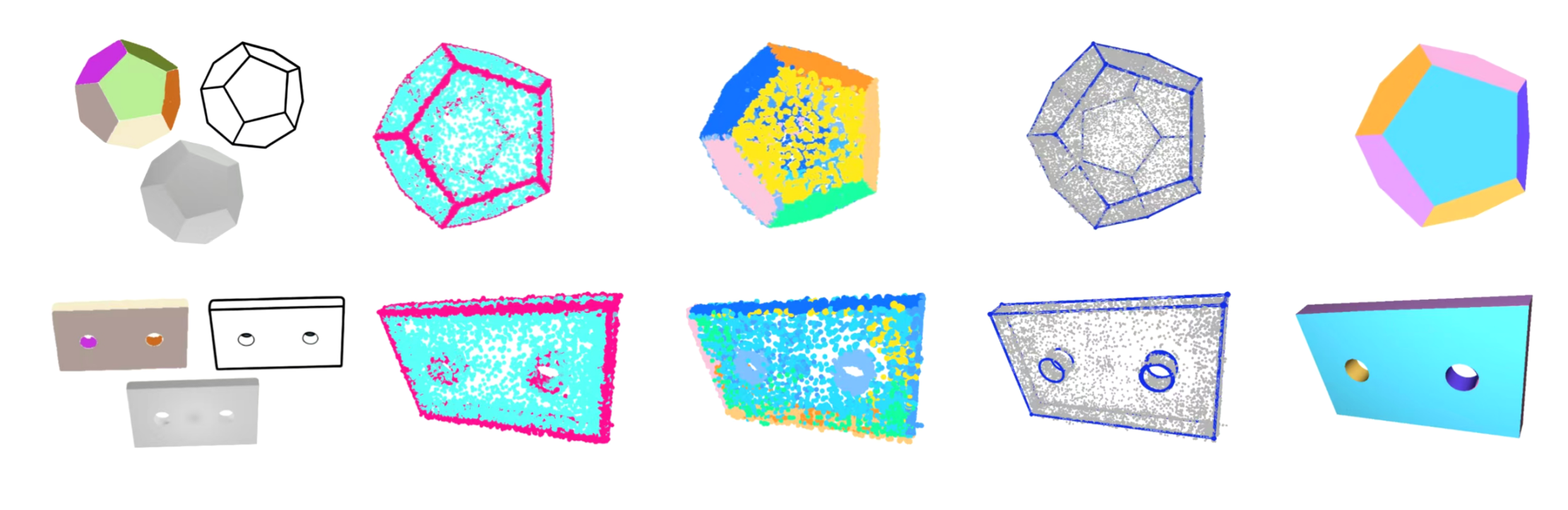}
    \put(19,0){\makebox[0pt][r]{{Multi-view images}}}
    \put(39,0){\makebox[0pt][r]{{Edge point clouds}}}
    \put(60,0){\makebox[0pt][r]{{Patch point clouds}}}
    \put(78,0){\makebox[0pt][r]{{Lines \& Curves}}}
    \put(96,0){\makebox[0pt][r]{{B-Rep Model}}}
    \end{overpic}
    \captionof{figure}{Given multi-view images, our pipeline reconstructs a CAD model through feature-aware Gaussian Splatting and parametric surface fitting.
    The Gaussian Splatting framework produces the geometric point cloud with edge and patch labels, which enables us to reconstruct parametric primitives from patch labels and extract lines and curves with the guidance of edge points.
}
    \label{fig:overview}
  \end{center}
}]
\footnotetext[2]{Corresponding author.}

\begin{abstract}
The boundary representation (B-Rep) models a 3D solid as its explicit boundaries: trimmed corners, edges, and faces. Recovering B-Rep representation from unstructured data is a challenging and valuable task of computer vision and graphics. Recent advances in deep learning have greatly improved the recovery of 3D shape geometry, but still depend on dense and clean point clouds and struggle to generalize to novel shapes. We propose B-Rep Gaussian Splatting (BrepGaussian), a novel framework that learns 3D parametric representations from 2D images. We employ a Gaussian Splatting renderer with learnable features, followed by a specific fitting strategy. To disentangle geometry reconstruction and feature learning, we introduce a two-stage learning framework that first captures geometry and edges and then refines patch features to achieve clean geometry and coherent instance representations. Extensive experiments demonstrate the superior performance of our approach to state-of-the-art methods.
\end{abstract}    
\section{Introduction}
\label{sec:intro}
The reconstruction of CAD models from acquired data has been a long-standing research problem. This process, often referred to as reverse engineering, aims to infer the underlying parametric and topological representations, not limited to geometric reconstruction.

Most research on CAD reconstruction has predominantly focused on point clouds. These methods take dense point clouds as input. Segmentation is applied to extract patch labels, where “patch” refers to regions bounded by the edges, followed by parametric reconstruction. 
Early learning-based approaches introduced supervised primitive segmentation networks that assign each point to a corresponding geometric primitive~\cite{Li2019SPFN,Deng2020Parsenet}. 

Subsequent works refined this paradigm by learning feature embeddings for point grouping instead of explicit primitive labels~\cite{Deng2020Parsenet}. More recent extensions combine edge and surface features to improve reconstruction quality and the consistency of fitted geometry~\cite{Wu2024SEDNet,qian2022pcernet}. 
Despite these advances, contemporary CAD reconstruction techniques are still constrained by two critical challenges: the expensive acquisition of high-quality point clouds and the heavy reliance on extensive manual annotation. In contrast, image data are far more accessible and scalable, yet there remains a large gap between image data and parametric 3D modeling.

Recently, multi-view 3D reconstruction has gained increasing attention with the emergence of neural rendering techniques such as NeRF~\cite{Mildenhall2020NeRF} and 3D Gaussian Splatting (3DGS)~\cite{Kerbl2023Gaussian}. These approaches enable high-fidelity geometry and appearance reconstruction directly from images. 
Motivated by these developments, we propose B-Rep Gaussian Splatting (BrepGaussian), a new framework designed to address this gap by extracting edge and patch features from multi-view images and directly applying them to 3D parametric reconstruction. 


The overall process of our BrepGaussian is as follows: We use image processing models to extract 2D edge and patch masks from multi-view images. We adopt 2D Gaussian Splatting (2DGS)~\cite{Wetzstein2024TwoDGS} as the fundational framework in our approach, where each Gaussian primitive is an oriented elliptical disk aligned to the surface. Each Gaussian contains additional edge and patch features. We employ two stage learning procedure: the first stage learns geometry and edge features, While in the second stage, we leverage the frozen geometric and edge features of the Gaussians learned in the first stage and use contrastive learning to obtain more complex patch instance labels.  Subsequently, we sample point clouds from the Gaussian primitives according to their elliptical shapes, ensuring that the sampled points are well aligned with the real surface distribution and assigned with corresponding labels. We employ a fitting model guided by the obtained labels to perform parametric reconstruction. 
Specifically, a constraint-guided primitive fitting module estimates parametric surfaces (planes, cylinders, and spheres) and progressively extracts edges and corners through geometric intersections. These elements are then assembled into a watertight B-Rep representation through constraint refinement and topological adjustment.

These steps constitute the entire pipeline for CAD reconstruction from multi-view images, as shown in Fig. \ref{fig:overview}. Our contributions are summarized as follows:
\begin{itemize}
    \item We introduce BrepGaussian, a novel framework for CAD reconstruction from multi-view images via 2D Gaussian Splatting and primitive fitting, providing highly accurate parametric reconstructions
    \item To improve segmentation quality, we design a two-stage learning process, where the second stage employs effective contrastive learning to tackle the challenges of complex patch learning
    \item We propose a constraint-guided primitive fitting module that effectively utilizes labels obtained from Gaussian training for accurate shape reconstruction.

\end{itemize}

To the best of our knowledge, BrepGaussian is the first framework that reconstructs complete B-Rep CAD models directly from multi-view images without any point cloud supervision.

\section{RELATED WORK}

\subsection{3D Parametric CAD Reconstruction}

3D parametric CAD reconstruction typically follows a divide-and-conquer paradigm, first analyzing the scene to assign semantic or geometric labels, then reconstructing each object by fitting parametric primitives and assembling them into structured, topology-consistent representations.

\noindent \textbf{Primitive fitting from point clouds.}
Early learning-based methods such as SPFN~\cite{Li2019SPFN} predict the type, parameters, and point-to-primitive assignments to reconstruct simple geometric elements. 
Later works, e.g., ParSeNet~\cite{Deng2020Parsenet}, extend this idea to more flexible surface fitting and richer parametric representations.
Subsequent works~\cite{Li2021CPFN,Liang2023HPNet,Qi2021Point2Cyl,Li2020GlobFit} adopt hybrid representation strategies to enhance primitive fitting, enabling more robust reconstruction in complex scenarios.

\noindent \textbf{Edge reconstruction.}
Edges are fundamental geometric features that define object boundaries and preserve structural details.
PIE-Net~\cite{Pham2020PIENet} first learns to infer parametric edge representations directly from point clouds.
DEF~\cite{Liu2020DEF} and EDC-Net~\cite{Zhang2021EDCNet} further explore neural extraction of sharp geometric features, while NerVE~\cite{Guo2024NerVE} achieves parametric curve extraction using neural volumetric edge fields.

\noindent \textbf{From primitives to CSG and B-Rep modeling.}
Primitive fitting provides local geometric parameters (planes, cylinders, Bézier surfaces), whereas Constructive Solid Geometry (CSG) and Boundary Representation (B-Rep) encode global topology and semantic hierarchy.
CSG-based approaches~\cite{Sharma2018CSGNet,Chen2020CAPRINet,Ji2020UCSGNet,Li2020CSGStump,Gao2023DualCSG} build shapes by combining parametric primitives through learnable boolean operations.
Recent methods such as SECAD-Net~\cite{Huang2024SECADNet} and SfmCAD~\cite{Chen2025SfmCAD} further learn sketch–extrude boolean sequences in a self-supervised manner.
Recent advances in B-Rep learning~\cite{Willis2021BRepNet,Zhang2023ComplexGen,Wu2024SEDNet,Zhu2024SplitFit,Zhou2024Point2CAD,qian2022pcernet}
have greatly improved CAD reconstruction.  
SED-Net~\cite{Wu2024SEDNet} combines surface and edge detection to improve instance segmentation and edge completeness. Similar method~\cite{qian2022pcernet} leverages surface patch segmentation to guide edge reconstruction, enabling more accurate and topologically consistent wireframe recovery. While Split-and-Fit~\cite{Zhu2024SplitFit} introduces a top-down paradigm via neural Voronoi partitioning and local primitive fitting.
Point2CAD~\cite{Zhou2024Point2CAD} refines surface fitting with continuity and intersection constraints.

However, existing methods heavily depend on high-quality point clouds and specialized network designs. Meanwhile, although recent generation methods\cite{Li2025CADDreamer,Alam2025GenCAD,Liu2025HoLa,You2025Img2CAD} have shown promising results in CAD generation, their performance remains limited for image-conditioned generation. Our BrepGaussian reconstructs CAD models from multi-view images via Gaussian Splatting, enabling accurate B-Rep generation without point cloud supervision.

\subsection{3D Reconstruction of Neural Rendering}

Image-based neural rendering techniques aim to recover 3D geometry from multi-view images by learning neural scene representations.

\noindent \textbf{Neural Radiance Fields.}
Early works such as Soft3D~\cite{Penner2017Soft3D} and NeRF~\cite{Mildenhall2020NeRF} employ volumetric rendering to synthesize novel views by integrating color and density along camera rays, while later extensions including Mip-NeRF~\cite{Barron2021MipNeRF} and NeuS~\cite{Wang2021NeuS} improve anti-aliasing and surface regularity.
Instant-NGP~\cite{Muller2022InstantNGP} further accelerates this process via hash-grid encoding for real-time optimization.


\noindent \textbf{3D Gaussian Splatting.}
Recently, 3DGS~\cite{Kerbl2023Gaussian} has emerged as a powerful explicit representation.
SuGaR~\cite{Wang2024SuGaR} aligns 3D Gaussians to surface normals for high-quality mesh reconstruction,
while 2DGS reduce 3D Gaussians to planar 2D Gaussian  primitives, improving surface accuracy.
Further developments enhance Gaussian representations with semantic and instance awareness, such as Segment Any 3D Gaussians~\cite{Wang2024SA3DGS}, Semantic Gaussians~\cite{Xu2024SemanticGS}, and Feature3DGS~\cite{Li2024Feature3DGS}.
Curve-Aware Gaussian Splatting~\cite{Zhang2025CurveAwareGS} and SGCR~\cite{Wang2024SGCR} recovers 3D parametric curves from multi-view images, 
while SketchSplat~\cite{Chen2024SketchSplat} explore sketch-based Gaussians for fine edge recovery and Hybrid Part-aware Representations~\cite{Liu2024HybridPartGS} develops part-level geometric reasoning.

Our BrepGaussian inherits the advantages of Gaussian Splatting and extends them to CAD reconstruction, enabling this representation to generalize beyond view synthesis toward structured 3D modeling tasks.

\begin{figure*}
    \centering
    \includegraphics[width=1\linewidth]{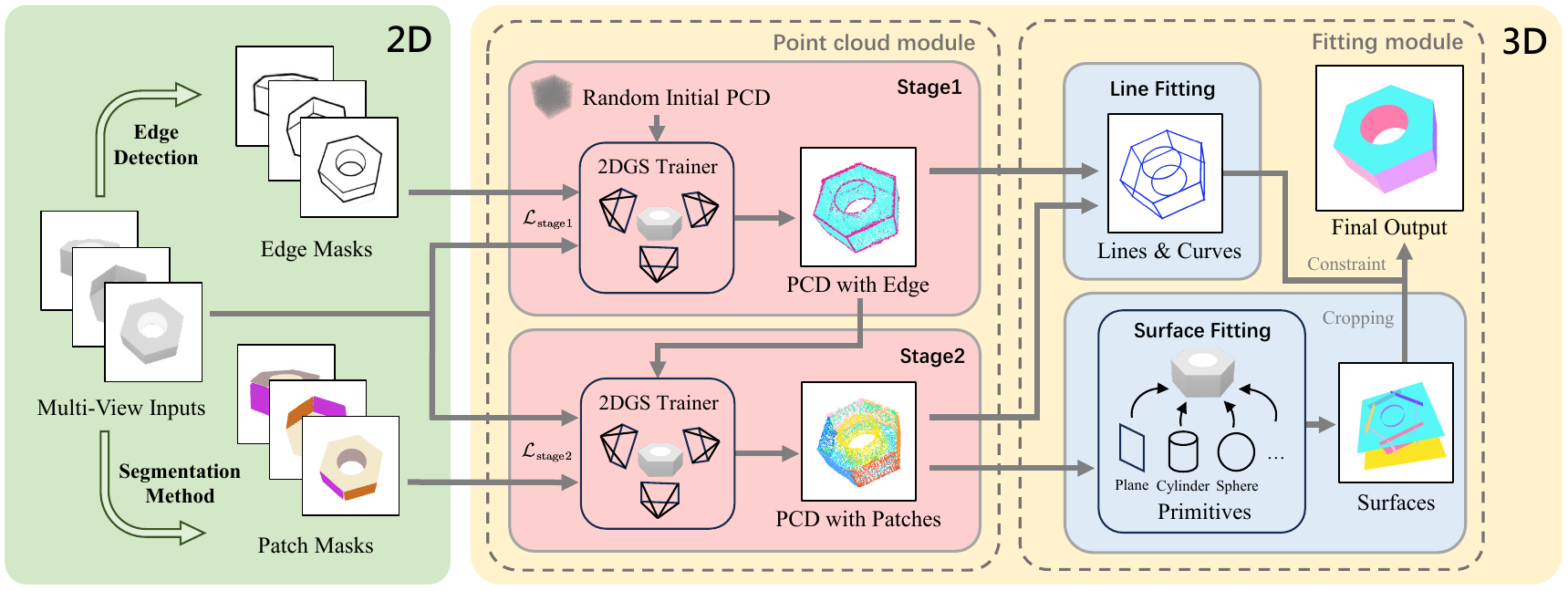}
    \caption{Overall pipeline of \textbf{BrepGaussian}.
Given multi-view RGB images of a CAD object, we extract edge and patch views using existing edge detection and segmentation models. 
These views drive a two-stage Gaussian Splatting model that 
predicts edge and patch labels on the reconstructed point cloud.
The fitted primitives are globally optimized to obtain the final  B-Rep model.}
    \label{fig:pipeline}
\end{figure*}

\section{METHOD}
Previous works for CAD reconstruction typically rely on high-quality point clouds, which are costly to acquire and require tedious annotations. With recent advances in image-based 3D reconstruction, we found it advantageous to extract 3D discrete features from multi-view images and reconstruct B-Rep from them. Overall,our method consists of the following steps, \emph{cf}.\ Fig.~\ref{fig:pipeline}:

\begin{enumerate}
\item Extract edge mask and per-patch mask from 2D images. We employ an edge detector and Segment Anything Model (SAM)~\cite{Kirillov2023_SAM} to extract 2D edge masks and per-patch masks from each view.
\item Use a 2DGS rasterizer with extra features in two stage reconstruction: First, we obtain a GS representation with geometry and clean edge semantics. For complex patch instances, we freeze all other Gaussian attributes and optimize corresponding embeddings with constrative learning. We convert the learned Gaussians into a point cloud.
\item Parametric fitting. we apply specific primitive types to recover parametric lines, curves and surfaces. Finally, we prune and refine spurious fragments to obtain clean CAD results.
\end{enumerate}

\subsection{Preliminary}
\textbf{3D Gaussian Splatting}~\cite{Kerbl2023Gaussian} represents a scene as a set of anisotropic 3D Gaussian primitives parameterized by position $\boldsymbol{\mu}_i\in\mathbb{R}^3$, covariance $\boldsymbol{\Sigma}_i\in\mathbb{R}^{3\times3}$, color $\mathbf{c}_i$, and opacity $\alpha_i$. 2DGS simplifies each primitive into a flat 2D Gaussian.
Each Gaussian is defined by its central point $\mathbf{p}_k$, two principal tangential vectors $\mathbf{t}_u, \mathbf{t}_v$, and scaling factors $\mathbf{s} = (s_u, s_v)$.
For a point $\mathbf{u}=(u,v)$ on the local tangent plane, the 2D Gaussian distribution is defined as
\begin{equation}
G(\mathbf{u}) = \exp\!\big(-\, \tfrac{1}{2}(u^2 + v^2)\big),
\end{equation}
and the projection onto screen space is rendered via weighted alpha blending:
\begin{equation}
{\mathbf{c}}(\mathbf{x}) =
\sum_{i=1} \mathbf{c}_i \alpha_i G_i(\mathbf{u(x)})
\prod_{j<i}\big(1 - \alpha_j G_j(\mathbf{u(x)}))\big).
\end{equation}

\noindent \textbf{Adaptivity of Gaussian Representations.}
Gaussian splatting has become a flexible framework that can adjust to different tasks.
Each Gaussian $g_i$ can be extended with a learnable feature embedding $\mathbf{f}_i \in \mathbb{R}^d$:
\begin{equation}
g_i = \{\, \boldsymbol{\mu}_i,\, \Sigma_i,\, \alpha_i,\, \mathbf{c}_i,\, \mathbf{f}_i \,\},
\end{equation}
Variants~\cite{Wang2024SA3DGS,Xu2024SemanticGS,Li2024Feature3DGS} leverage $\mathbf{f}_i$ for segmentation learning.
Curve-Aware Gaussian Splatting~\cite{Zhang2025CurveAwareGS} extends Gaussian splatting to parametric curve reconstruction. 
Each curve $\mathcal{C}(t)$ is represented as a cubic Bézier curve:
\begin{equation}
\boldsymbol{\mu}(t) = 
\mathbf{P}_0 (1-t)^3 + 3\mathbf{P}_1 t(1-t)^2 
+ 3\mathbf{P}_2 t^2(1-t) + \mathbf{P}_3 t^3,
\end{equation}
where $\{\mathbf{P}_k\}_{k=0}^3$ are control points.
A sequence of Gaussians is sampled along $\mathcal{C}(t)$. This formulation yields differentiable curve fitting.

Building on 2DGS, our method extends each Gaussian with learnable features. Since B-Rep reconstruction is inherently more complex than curve fitting, we adopt a two-stage reconstruction strategy that infers patch segmentation from Gaussian Splatting and fits parametric surfaces for CAD recovery.

\subsection{Learning to Instantiate Edges and Patches}
In the context of the overall CAD reconstruction pipeline, this step consists of four parts: (1) 2D label extraction --- edges and per-patch masks for each view; (2) Stage~1 training --- learning 3D geometry and edge semantics; (3) Stage~2 training --- learning 3D patch instances; (4) conversion from Gaussian to point cloud --- sampling Gaussians to obtain a clean, labeled point cloud for downstream fitting.

Given multi-view images of a CAD part, we first extract 2D labels per view. we use an edge detector to obtain edge mask $E$. The extracted edge mask serves as prompts to SAM to generate patch mask.

We adopt 2DGS as our rendering backbone. Its representation carries no artificial thickness, aligning well with CAD geometry dominated by planes and low-curvature surfaces. In our experiments, we found that a \emph{two-stage training strategy} yields the
most stable and accurate results. In Stage~1, we jointly learn the
geometry of CAD objects and their edge semantics. In Stage~2, we
freeze the geometric parameters (e.g., positions $x,y,z$ and spherical harmonics) so that the subsequent complex training of patch instances does not damage the reconstructed geometry, allowing the network to focus on a single, well-defined learning task.

In Stage~1, once we obtain the edge map $E$, we assign each Gaussian $g_i$ with a
single scalar edge value $e_i\in[0,1]$, which suffices to encode edge semantics.
During splatting, the edge value is rendered analogously to RGB using the same
alpha-compositing weights: 
\begin{equation}
w_i \;=\; \alpha_i \prod_{j<i} (1-\alpha_j) \qquad
E(u) \;=\; \sum_{i=1}^N w_i\, e_i
\end{equation}
where $\alpha_i$ is the opacity of the $i$-th Gaussian, $w_i$ is its accumulated
visibility weight along the viewing ray, and $e_i$ is the learnable edge
probability. The rendered edge map $E(u)$ is therefore
an alpha-weighted accumulation of per-Gaussian edge values.
In Stage~2, we train 3D instance labels of patches
based on the obtained 2D patch masks. Each Gaussian $g_i$ is assigned with a high-dimensional feature vector
$\mathbf{f}_i\!\in\!\mathbb{R}^d$, as a single scalar is insufficient for feature learning. Since masks from different views lack consistency, we adopt an effective contrastive  learning strategy to guide feature learning correctly.
Each feature channel is rendered similarly to the edge value in Stage~1.

To convert Gaussians into a point cloud, we sample from each Gaussian. 
In our experiments, we found that regions near edges required a large number of elongated Gaussians, whereas mostly flat regions required only a few nearly spherical Gaussians. 
Accordingly, we convert each Gaussian into a point set by sampling its center, and for ellipses whose major axis is not extremely larger than the minor axis, we sample four additional points along the ellipse to closely approximate the real surface.

\subsection{Loss Functions}

\textbf{Stage~1 Loss.} We design a composite loss that jointly
supervises geometry reconstruction and edge prediction.The geometry in our framework is optimized following the original formulation of Gaussian Splatting:
\begin{equation}
\mathcal{L}_{\text{geo}}
= (1-\lambda)\,\mathcal{L}_1
+ \lambda\mathcal{L}_{D-SSIM}
\end{equation}
where $L_1$ and $\mathcal{L}_{D-SSIM}$ are supervised by the original image.


We further introduce an edge-aware loss to determine whether a Gaussian lies on object boundaries:

\begin{equation}
\mathcal{L}_{\text{edge}} = \sum_{i\in E}\|I_i-\hat{I_i}\|_{2}^2.
\end{equation}

The final objective is defined as:
\begin{equation}
\mathcal{L}_{\text{stage1}}
= \mathcal{L}_{\text{geo}} + 0.1\,\mathcal{L}_{\text{edge}}.
\end{equation}

\noindent \textbf{Stage~2 Loss} is designed to aggregate Gaussian features within the same mask while repelling those across different masks with triplet loss. We measure pairwise distances between feature samples using cosine distance:
\begin{equation}
d(\mathbf{p}_i,\mathbf{p}_j)
=1-\tilde{\mathbf{f}}_{\mathbf{p}_i}\cdot
\tilde{\mathbf{f}}_{\mathbf{p}_j}
\end{equation}
where $\tilde{\mathbf{f}}_{\mathbf{p}}$ denotes the $\ell_2$-normalized feature vector
at pixel $\mathbf{p}$, obtained by opacity-weighted accumulation of all Gaussian features.

Let $\mathcal{M}_k$ denote the $k$-th 2D mask region, For each valid mask $\mathcal{M}_k$, 
we randomly sample an anchor point 
$\mathbf{p}_a\!\in\!\mathcal{M}_k$, 
a positive $\mathbf{p}_p\!\in\!\mathcal{M}_k\setminus\{\mathbf{p}_a\}$,
and a set of negatives 
$\mathcal{C}_k=\!\!\bigcup_{k'\neq k}\mathrm{sample}(\mathcal{M}_{k'})$. The hardest negative sample $\mathbf{p}_n
=\arg\min_{\mathbf{p}\in\mathcal{C}_k} 
d(\mathbf{p}_a,\mathbf{p})$ is selected from the candidate set $\mathcal{C}_k$ as the closest feature from other samples.

The overall triplet loss is then defined as
\begin{equation}
\mathcal{L}_{\text{tri}}
=\frac{1}{|\mathcal{T}|}
\sum_{(\mathbf{p}_a,\mathbf{p}_p,\mathbf{p}_n)\in\mathcal{T}}
\max\!\big(0,\,
d(\mathbf{p}_a,\mathbf{p}_p)
-d(\mathbf{p}_a,\mathbf{p}_n)
+m\big),
\end{equation}
where $\mathcal{T}$ is the set of sampled triplets and $m$ is the margin
hyperparameter. This enforces compact features within each patch and clear separation
across different patches.

\begin{figure}[t]
	\centering
	\includegraphics[width=0.6\linewidth]{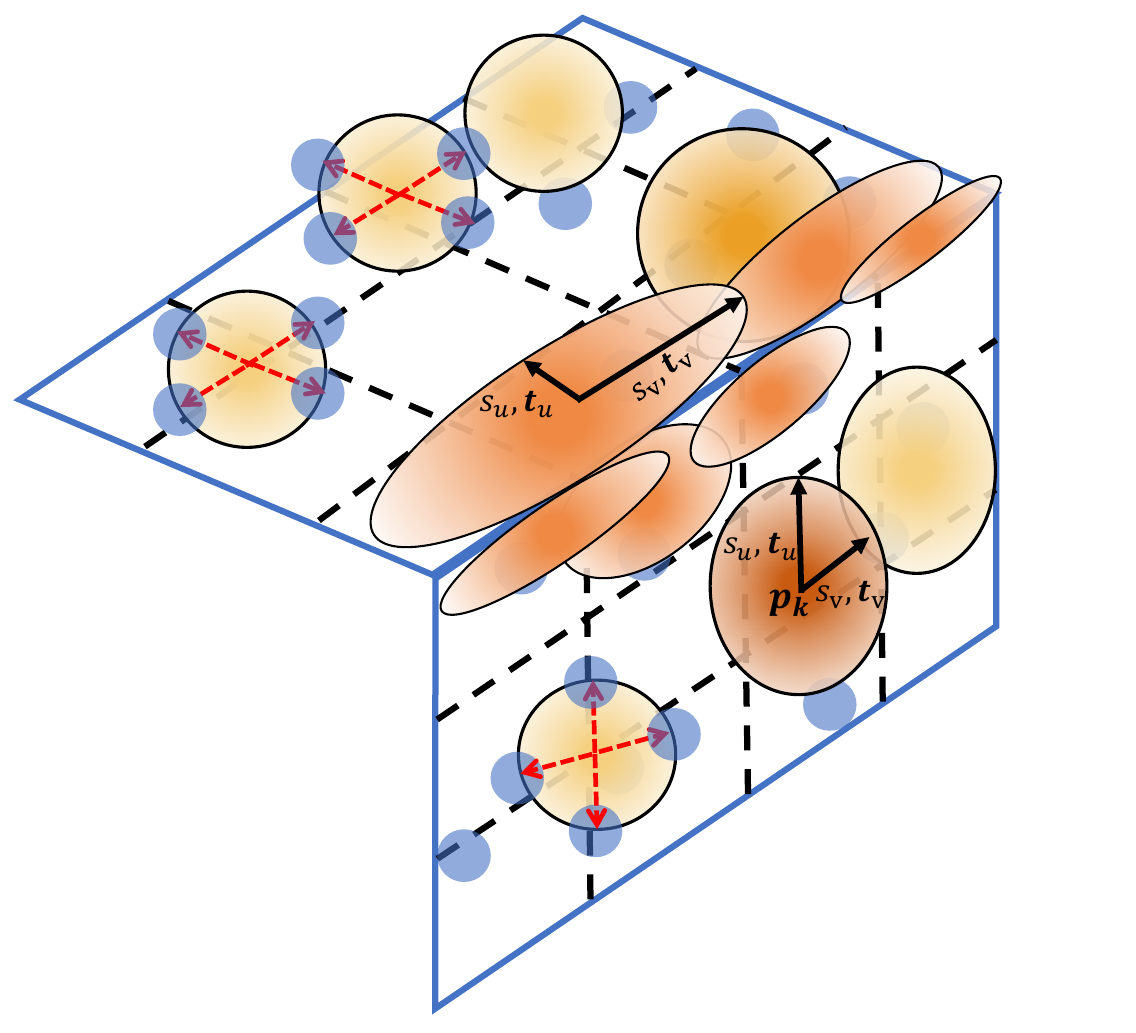}
    \caption{Illustration of optimized 2D Gaussians. flat regions use fewer nearly spherical Gaussians, while edge regions require more elongated Gaussians.}
    \vspace{-1.5em}
\end{figure}

\subsection{B-Rep fitting}
We design a constraint-guided primitive fitting module that fits planes, cylinders, and spheres according to the predicted patch labels, and assembles them into a watertight B-Rep representation.
Overall, our process consists of primitive fitting, hierarchical extraction from surfaces to edges and corners, point cloud constraint refinement, bottom-up assembly from corners to edges and surfaces, and a final topological adjustment to obtain a complete B-Rep representation.

For each patch, we define three primitive models: plane, cylinder, and sphere, formulated as parametric surfaces for subsequent geometric fitting and optimization.
To achieve robust parameter estimation, we adopt the RANSAC\cite{Fischler1981RANSAC} paradigm on each primitive model, allowing stable geometric fitting even with noisy point cloud.
In principle, RANSAC iteratively samples minimal point subsets, fits candidate models, and evaluates their inlier ratios based on geometric distance metrics to identify the model with the highest consensus.
Next, intersections between primitive pairs are computed to yield corresponding lines and curves in geometric space. 
These results are then constrained by edge point clouds to extract valid line and curve segments, where edge points are projected onto the line and curve to determine their parameter ranges $t$.
Candidate corner points are obtained from the intersections of three planes or two lines. These candidates are clustered to generate final corner points.
Finally, each surface is refined under the constraints of line and curve segments, while Boolean operations are employed to ensure a clean and watertight B-Rep reconstruction.
More algorithm details are provided in the supplementary material.


\begin{figure*}[t]
  \centering
  \includegraphics[width=1\linewidth]{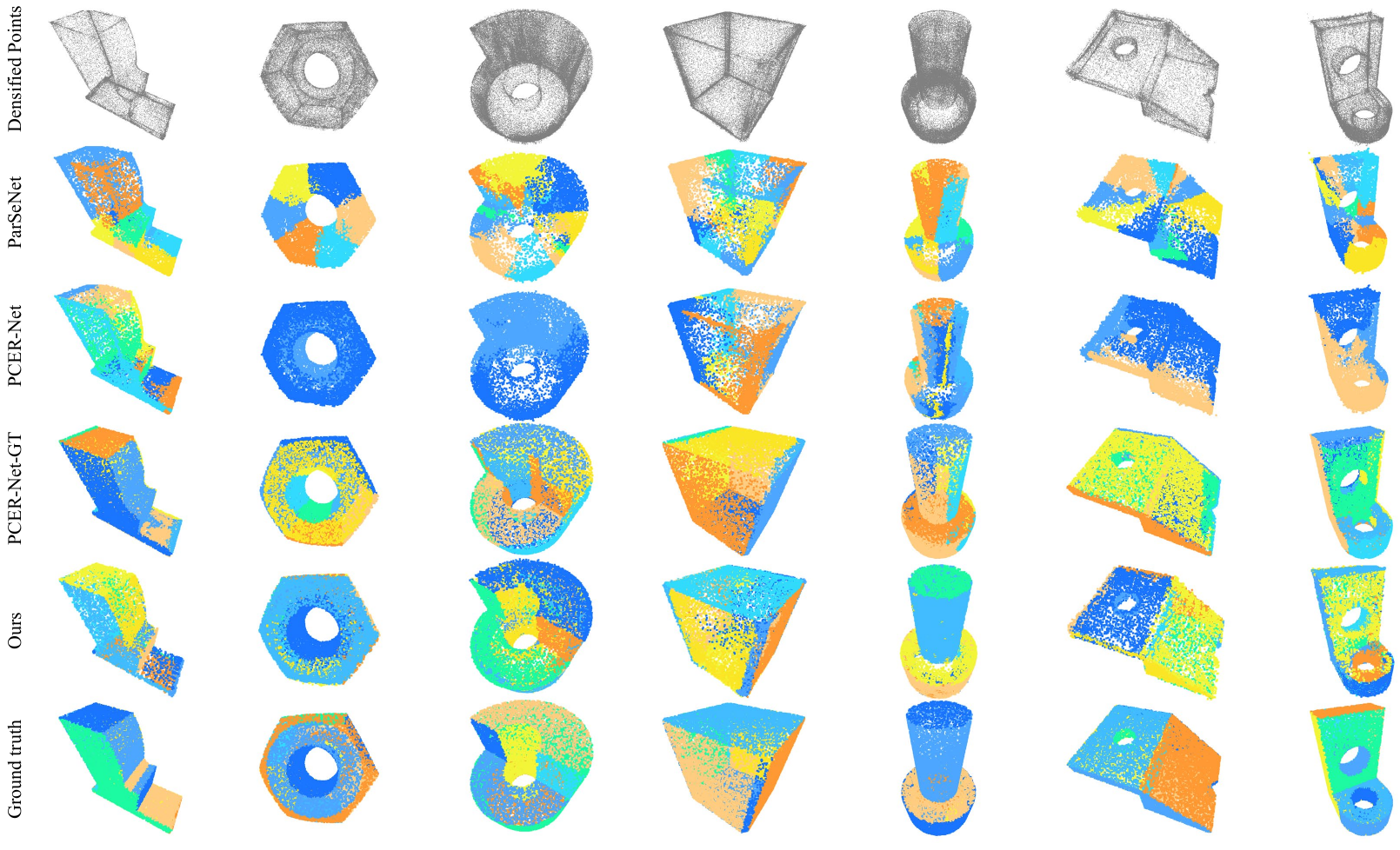}
  \caption{Qualitative comparison on patch segmentation. Our BrepGaussian produces cleaner and more consistent patch segmentation.}
  \label{fig:experiment1}
\end{figure*}



\begin{figure*}[t]
  \centering
  \includegraphics[width=1\linewidth]{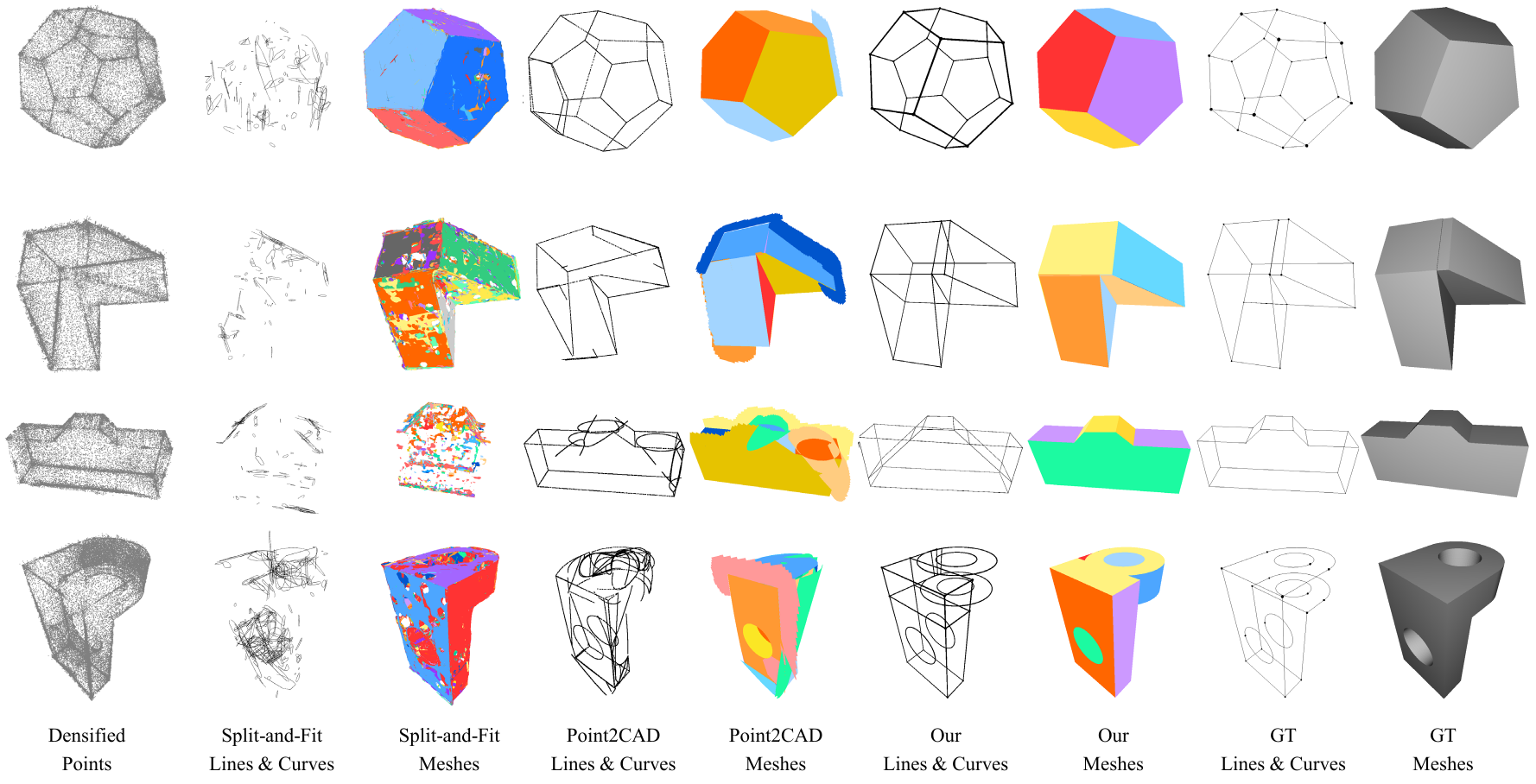}
  \caption{Qualitative comparison on CAD reconstruction. Our BrepGaussian exhibit more accurate geometry.}
  \label{fig:experiment2}
\end{figure*}

\section{EXPERIMENTS}
\subsection{Implementation Details}
We evaluate the BrepGaussian on the ABC-NEF~\cite{Ye_2023_CVPR} subset of the ABC Dataset. The ABC Dataset~\cite{Koch_2019_CVPR} comprises over one million CAD models with explicitly parameterized curves and surfaces, and ABC-NEF is a challenging subset of this collection that offers 50 multi-view images for each object. We employ SAM~\cite{Kirillov2023_SAM} to generate per-patch mask image for each input view, in which multiple masks are associated with the individual patches of the object. Since the ABC-NEF images are low-texture, using SAM directly produces inconsistent and fragmented masks across views. We refined each object manually using a script, spending around three minutes per object.
In training, Stage~2 uses a patch feature of dimension 16.

\subsection{Comparisons}
\textbf{Settings.} 
We extract per-patch labels, curve labels and uniformly sampled point cloud (20k points) from the ABC dataset as ground truth. We evaluate our method against several baseline methods in segmentation and CAD reconstruction from point cloud. All these methods are supervised on the ABC dataset and thus exhibit strong dataset dependency.

In segmentation, for fair comparison, we use point clouds reconstructed from multi-view images as inputs to baselines. These point clouds, generated from our Gaussian Splatting and resampled through our up-sampling procedure, are dense and well aligned with the ground-truth point clouds. (see Fig.\ref{fig:experiment1}, first row, abbreviated as Densified Points). We also include several baselines that take ground-truth point clouds as input; however, their results are not directly comparable to ours due to the higher input quality.
In patch segmentation, we compare our results with ParSeNet~\cite{Deng2020Parsenet}, HPNet~\cite{Liang2023HPNet}, PCER-Net~\cite{qian2022pcernet}, and SED-Net~\cite{Wu2024SEDNet}.
In edge segmentation, we compare our results with PCER-Net~\cite{qian2022pcernet} and SED-Net~\cite{Wu2024SEDNet}. 
SED-Net and HPNet fails to generalize on densified point clouds. Therefore, both methods are not included in the evaluation on densified point clouds. 
Unlike previous methods that predict a fixed set of semantic classes, our BrepGaussian outputs continuous patch values learned through contrastive objectives.
The continuous labels are then refined through automatic filtering and label merging to obtain clean, discrete patches.

In CAD reconstruction, we compare our results with Point2CAD~\cite{Zhou2024Point2CAD} and Split-and-Fit~\cite{Zhu2024SplitFit}. Point2CAD is only a fitting-based method. To ensure fairness, we run it with our predicted labels and with labels from PCER-Net. For Split-and-Fit, we use our densified point clouds.

To quantitatively evaluate the quality of segmentation, we compute Precision and Recall based on geometric matching between predicted
and ground truth patches. 
Let $\mathcal{S}=\{S_i\}_{i=1}^{N_s}$ denote the set of predicted surface patches
and $\mathcal{G}=\{G_j\}_{j=1}^{N_g}$ denote the ground-truth patches.
Each patch $S_i$ or $G_j$ is represented by a set of 3D points.
For every predicted patch $S_i$, we measure its mean minimal distance to each ground-truth
patch $G_j$ as
\begin{equation}
    D(S_i, G_j) = \frac{1}{|S_i|}\sum_{\mathbf{p}\in S_i} 
    \min_{\mathbf{q}\in G_j}\|\mathbf{p}-\mathbf{q}\|_2.
\end{equation}
A predicted patch $S_i$ is regarded as a correct match if 
$\min_j D(S_i,G_j) \le \tau$, where $\tau=0.08$(unit length). 
The patch-level precision is defined as:
\begin{equation}
    \mathrm{Prec} =
    \frac{1}{N_s}|\{S_i \in \mathcal{S} \mid \min_j D(S_i,G_j) \le \tau\}|.
\end{equation}
Similarly, for each ground-truth patch $G_j$, We denote the recall:
\begin{equation}
    \mathrm{Rec} =
    \frac{1}{N_g}|\{G_j \in \mathcal{G} \mid \min_i D(G_j,S_i) \le \tau\}|.
\end{equation}
Both metrics offer an incomplete assessment of segmentation quality. 
When the predicted patches are relatively small, Precision tends to increase. While predicting overly large patches leads to higher Recall. We also report the F$_1$ score, which is the harmonic mean of Precision and Recall:
\begin{equation}
    \mathrm{F_1} =
    \frac{2 \cdot \mathrm{Prec} \cdot \mathrm{Rec}}
         {\mathrm{Prec} + \mathrm{Rec}}.
\end{equation}

\noindent \textbf{Quantitative Comparison.}
Table~\ref{tab:seg_patch} and~\ref{tab:seg_edge} reports the quantitative comparison for patch segmentation and edge segmentation.
HPNet produces continuously distributed and overly fragmented labels.
This fragmentation inflates the Precision to 1.0 but result in very low Recall, and prevents us from automatically extracting valid labels for coherent CAD reconstruction. Consequently, HPNet fails to achieve state-of-the-art performance.
Based on the densified point clouds reconstructed from multi-view inputs, our method outperforms existing approaches.
Compared with methods that take ground-truth point clouds as input, 
our results are better than PCER-Net but lower than SED-Net.

We use Chamfer Distance (CD) and Hausdorff Distance (HD) to evaluate the reconstructed surfaces and edges.
As shown in Tab.~\ref{tab:cad_reconstruction}, our method achieves the best performance on curve reconstruction and slightly lower scores on surface metrics compared with Point2CAD (with our labels).
Nevertheless, the qualitative analysis indicates that its higher scores are due to redundant patch reconstruction, leading to inferior reconstruction quality.

\begin{table}[t]
\centering
\caption{Quantitative comparison of different methods for patch segmentation. The
best results are shown in bold.}
\label{tab:seg_patch}
\setlength{\tabcolsep}{3pt}  
\small                       
\begin{tabular}{lcccc}
\toprule
\textbf{Method} & \textbf{Input} & \textbf{Prec~$\uparrow$} & \textbf{Rec~$\uparrow$} & \textbf{F$_1$~$\uparrow$} \\
\midrule
ParSeNet~\cite{Deng2020Parsenet}  & GT Points       & 0.5108 & 0.2646 & 0.3486 \\
PCER-Net~\cite{qian2022pcernet}  & GT Points      & 0.8760 & 0.9119 & 0.8936 \\
SED-Net~\cite{Wu2024SEDNet}   & GT Points    & 0.9490 & \textbf{1.0000} & \textbf{0.9738} \\
HPNet~\cite{Liang2023HPNet}     & GT  Points      & \textbf{1.0000} & 0.2142 & 0.3528 \\
\midrule
ParSeNet~\cite{Deng2020Parsenet}  & Densified  & 0.6229 & 0.2364 & 0.3427 \\
PCER-Net~\cite{qian2022pcernet}  & Densified  & 0.5357 & 0.7924 & 0.6392 \\
Ours     & multi-view & \textbf{0.8903} & \textbf{0.9181} & \textbf{0.9040} \\

\bottomrule
\end{tabular}

\vspace{1em}

\centering
\caption{Quantitative comparison of different methods for edge segmentation performance.}
\label{tab:seg_edge}
\setlength{\tabcolsep}{3pt}  
\small                        
\begin{tabular}{lcccc}
\toprule
\textbf{Method} & \textbf{Input} & \textbf{Prec~$\uparrow$} & \textbf{Rec~$\uparrow$} & \textbf{F$_1$~$\uparrow$} \\
\midrule
PCER-Net~\cite{qian2022pcernet}  & GT Points       & 0.8807    & 0.9563   & 0.9169 \\
SED-Net~\cite{Wu2024SEDNet}   & GT Points       & \textbf{0.9693} & \textbf{0.9913} & \textbf{0.9802} \\
\midrule
PCER-Net~\cite{qian2022pcernet}  & Densified & 0.7149    & 0.8349    & 0.7703 \\
Ours    & multi-view & \textbf{0.9350} & \textbf{0.9253} & \textbf{0.9301} \\

\bottomrule
\end{tabular}
\end{table}

\noindent \textbf{Qualitative Comparison.}
We illustrate the qualitative comparisons for patch segmentation in Fig.~\ref{fig:experiment1}. Suffix “-GT” indicates results tested on ground-truth point clouds.
As shown in the visualization, our method produces the most accurate and visually coherent patch segmentation, with clear patch boundaries. Among other methods, PCER-Net-GT performs second best. It over-segments the cylinder of the fifth object and ignores the intersection between planes on the sixth object. Other methods based on densified points input yield inferior segmentation performance.

We further present qualitative comparisons on CAD reconstruction in Fig.~\ref{fig:experiment2}.
Our method produces the cleanest and most compact results with high geometric quality.
In contrast, Point2CAD tends to predict redundant patches, it also struggles with slightly more complex shapes, often generating excessive and fragmented surfaces.
Split-and-Fit results in fragmented patches with distortion and breakage.

\begin{table}[t]
\centering
\caption{Quantitative comparison of different methods for CAD reconstruction.}
\label{tab:cad_reconstruction}
\setlength{\tabcolsep}{3pt}  
\small
\begin{tabular}{l|c|cc|cc}
\toprule
        \multirow{2}{*}{\textbf{Method} }  & \textbf{Input} & \multicolumn{2}{c|}{$\mathbf{D_c (10^{-2}) \downarrow}$} & \multicolumn{2}{c}{$\mathbf{D_h (10^{-1}) \downarrow}$} \\
               & \textbf{Points} & Surface & Curve & Surface & Curve   \\ 
    \midrule
    Point2CAD~\cite{Zhou2024Point2CAD} & Our labels & \textbf{3.38} & 5.42 & \textbf{2.413} & 3.858 \\
    Point2CAD~\cite{Zhou2024Point2CAD} &  PCER-Net & 7.08 & 20.45 & 3.394 & 7.276 \\
    Split-and-Fit~\cite{Zhu2024SplitFit} & Densified & 6.23 & 13.98 & 3.523  & 4.962\\
    Ours & Our points & 4.90 & \textbf{5.01}  & 3.351 & \textbf{3.626}\\
        
\bottomrule
\end{tabular}
\vspace{-2mm}
\end{table}

\subsection{Ablation Studies}
We conduct ablation studies to evaluate the effectiveness of each module, including the two-stage training scheme, contrastive learning loss, boundary extraction, and densification.
w/o two-stage denotes using only single-stage training, and w/o triplet loss replaces the triplet loss with a pairwise contrastive loss.
The results clearly demonstrate that each module contributes to the overall performance, and removing any component leads to a noticeable degradation, confirming the effectiveness of our design. As shown in Tab.~\ref{tab:ablation_seg} and Tab.~\ref{tab:cad_comparison}, removing any component results in a clear performance drop, confirming that each module plays an important role in our framework.
Meanwhile, we fine-tuned SAM to show that the pipeline can be made fully automatic with a minor loss in accuracy, as shown in Tab.~\ref{tab:manual}.

\begin{table}[ht]
\centering
\caption{Ablation study on patch segmentation.}
\label{tab:ablation_seg}
\setlength{\tabcolsep}{5pt}
\small
   \vspace{-2mm}

\begin{tabular}{lccc}

\toprule
\textbf{Pipeline} & \textbf{Prec~$\uparrow$} & \textbf{Rec~$\uparrow$} & \textbf{F1~$\uparrow$} \\
\midrule
w/o two-stage learning & 0.8709 & 0.6805 & 0.7672 \\
w/o triplet loss $\mathcal{L}_{tri}$ & 0.8395 & \textbf{0.9374} & 0.8857 \\
Full Model & \textbf{0.8903} & 0.9181 & \textbf{0.9040} \\ 
\bottomrule
\vspace{0.5em}
\end{tabular}

\centering
\vspace{-2mm}
\caption{Ablation study on CAD reconstruction.}
\label{tab:cad_comparison}
\setlength{\tabcolsep}{3pt}  
\small
   \vspace{-2mm}

\begin{tabular}{l|cc|cc}
    \toprule
    \multirow{2}{*}{\textbf{Pipeline}} & \multicolumn{2}{c|}{$\mathbf{D_c (10^{-2}) \downarrow}$} & \multicolumn{2}{c}{$\mathbf{D_h (10^{-1}) \downarrow}$} \\
           & Surface & Curve & Surface & Curve   \\ 
    \midrule
    w/o edge segmentation & \textbf{4.63} & 5.83 & 3.601 & 3.705  \\
    w/o densification & 5.99 & $\backslash$ & 4.184 & $\backslash$  \\
    Full Model & 4.90 & \textbf{5.01} & \textbf{3.351} & \textbf{3.626} \\ 
    \bottomrule
    
\end{tabular}
\vspace{-5.5mm}
\end{table}

\begin{table}[h]
  \centering
  \caption{Ablation study on mask quality.}
\label{tab:manual}
   \vspace{-2mm}
\resizebox{\linewidth}{!}{\renewcommand{\arraystretch}{0.8}
  \begin{tabular}{c|ccc|c}
    \toprule
    \textbf{Data Preparation}  & $\mathbf{Prec\uparrow}$ & $\mathbf{Rec\uparrow}$ & $\mathbf{F_1\uparrow}$ & $\mathbf{D_c (10^{-2}) \downarrow}$ \\
    \midrule
    Manual Correction & 0.8704 & \textbf{0.9191} & \textbf{0.8941} & \textbf{5.04} \\
    Fine-tuned SAM & \textbf{0.8896} & 0.8363 & 0.8621 & 5.33 \\
    Vanilla SAM & 0.8435 & 0.8090 & 0.8259 & 5.66 \\
    \midrule
  \end{tabular}
}
\end{table}

\begin{figure}[ht]
  \centering
  \includegraphics[width=0.5\textwidth]{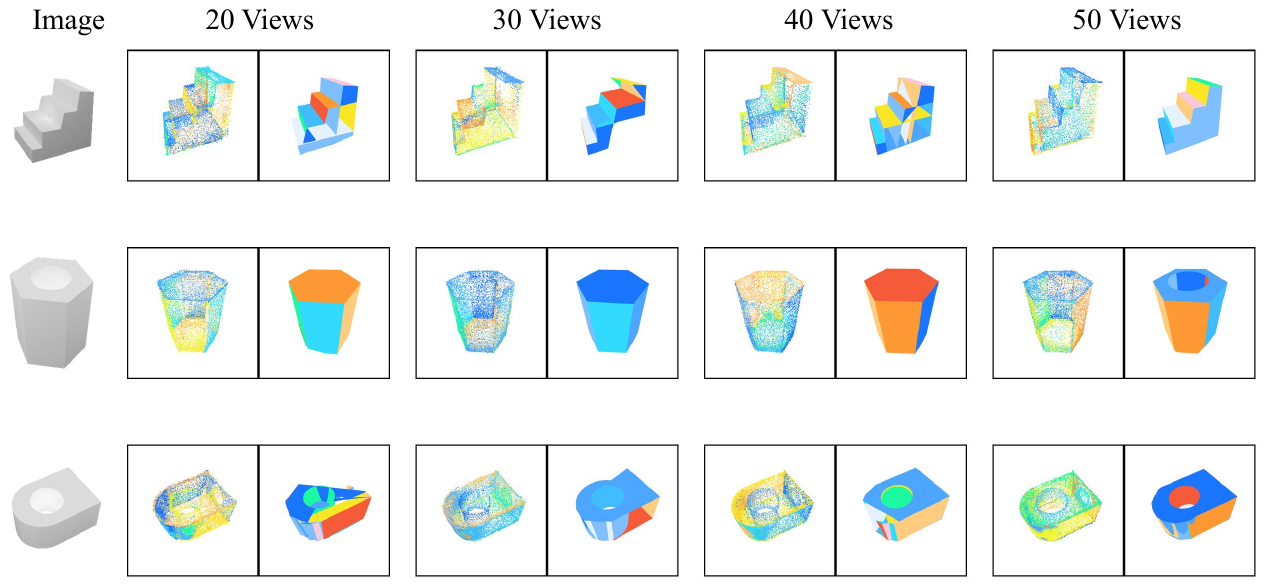}
  \caption{From left to right, we present 2D images, point cloud and final B-Rep in 20, 30, 40 and 50 views.}
  \label{fig:multi-view}
  \vspace{-2mm}
\end{figure}

We further investigate the effect of the number of input views on reconstruction quality.
Models are trained with 20, 30, 40, and 50 views, respectively.
As illustrated in Fig.~\ref{fig:multi-view}, using 30 views or fewer results in noisy patch labels and incomplete reconstructions, while 50 views are required to produce clean, dense, and high-quality B-Rep reconstructions.


\subsection{Real-World Scene}
\begin{figure}[h]
  \begin{overpic}[width=0.5\textwidth]{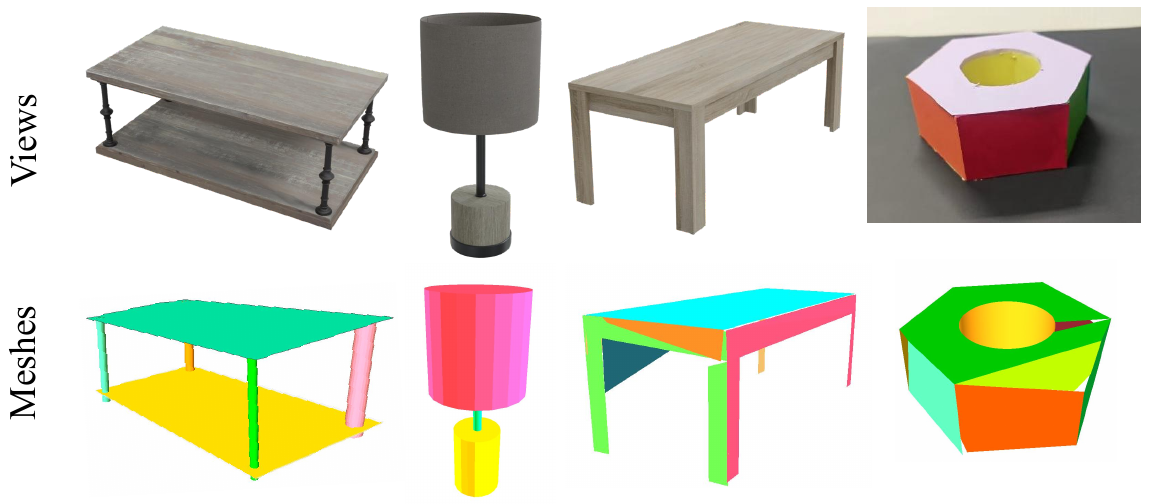}
  \end{overpic}
  \vspace{-1em}
  \caption{Experiments on real-world scenes from ABO dataset~\cite{collins2022abo} and smartphone photos.}
  \label{fig:realworld}
  \vspace{-2mm}
\end{figure}

We further evaluate our method on real-world scenes from the ABO dataset~\cite{collins2022abo} and smartphone photos.
As shown in Fig.~\ref{fig:realworld}, our BrepGaussian again produces clean CAD models, demonstrating strong generalization to real-world data.

\subsection{Limitations}
We admit that the quality of points and labels from multi-views limits our capabilitiy to complex objects. The overall pipeline is still relatively complex.
\section{Conclusions}

We introduced BrepGaussian, a novel framework for 3D CAD reconstruction directly from multi-view images.
Our key insight is to map 2D information into 3D Gaussian features that support parametric fitting of geometric primitives. 
The proposed two-stage Gaussian learning, with specific fitting and topological optimization, enables accurate and consistent B-Rep modeling.
To the best of our knowledge, BrepGaussian is the first approach to achieve complete CAD reconstruction from images without any point cloud supervision.
When compared with methods that take point clouds reconstructed from the same multi-view images as input, our framework outperforms them.
These results demonstrate the promise and feasibility of recovering structured 3D geometry purely from 2D images.

\section*{Acknowledgments}
This work was supported in part by the Fundamental and Interdisciplinary Disciplines Breakthrough Plan of the Ministry of Education of China (No. JYB2025XDXM118), and the Natural Science Foundation of Jiangsu Province (No. BK20251195), and Collaborative Innovation Center of Novel Software Technology and Industrialization.

{
    \small
    \bibliographystyle{ieeenat_fullname}
    \bibliography{main}

@String(CVPR= {IEEE Conf. Comput. Vis. Pattern Recog.})

@String(ICCV= {Int. Conf. Comput. Vis.})

@String(ECCV= {Eur. Conf. Comput. Vis.})

@String(NIPS= {Adv. Neural Inform. Process. Syst.})

@String(TOG= {ACM Trans. Graph.})

@String(AAAI = {AAAI})

@String(SIGGRAPH = {ACM SIGGRAPH Annual Conference})

@String(APS = {Applied Sciences})

@String(TMLR = {Trans. Mach. Learn. Res.})

@inproceedings{Li2019SPFN,
  author = {Li, Lingxiao and Sung, Minhyuk and Dubrovina, Anastasia and Yi, Li and Guibas, Leonidas},
  title = {Supervised Fitting of Geometric Primitives to 3D Point Clouds},
  booktitle = CVPR,
  year = {2019}
}

@inproceedings{Deng2020Parsenet,
  author = {Sharma, Gopal and Liu, Difan and Maji, Subhransu and Kalogerakis, Evangelos and Chaudhuri, Siddhartha and Měch, Radom{\'\i}r},
  title = {ParSeNet: A Parametric Surface Fitting Network for 3D Point Clouds},
  booktitle = ECCV,
  year = {2020}
}

@inproceedings{Li2021CPFN,
  author = {L{\^e}, Eric-Tuan and Sung, Minhyuk and Ceylan, Duygu and M{\v{e}}ch, Radom{\'\i}r and Boubekeur, Tamy and Mitra, Niloy J.},
  title = {CPFN: Cascaded Primitive Fitting Networks for High-Resolution Point Clouds},
  booktitle = ICCV,
  year = {2021}
}

@inproceedings{Liang2023HPNet,
  author = {Yan, Siming and Yang, Zhenpei and Ma, Chongyang and Huang, Haibin and Vouga, Etienne and Huang, Qixing },
  title = {HPNet: Deep Primitive Segmentation Using Hybrid Representations},
  booktitle = ICCV,
  year = {2021}
}

@inproceedings{Qi2021Point2Cyl,
  author = {Uy, Mikaela Angelina and Chang, Yen-Yu and Sung, Minhyuk and Goel, Purvi and Lambourne, Joseph and Birdal, Tolga and Guibas, Leonidas},
  title = {Point2Cyl: Reverse Engineering 3D Objects from Point Clouds to Extrusion Cylinders},
  booktitle = CVPR,
  year = {2022}
}

@article{Li2020GlobFit,
  author = {Li, Yangyan and Wu, Xiaokun and Chrysathou, Yiorgos and Sharf, Andrei and Cohen-Or, Daniel and Mitra, Niloy J.},
  title = {GlobFit: Consistently Fitting Primitives by Discovering Global Relations},
  journal = TOG,
  year = {2011}
}

@inproceedings{Pham2020PIENet,
  author = {Wang, Xiaogang and Xu, Yuelang and Xu, Kai and Tagliasacchi, Andrea and Zhou, Bin and Mahdavi-Amiri, Ali and Zhang, Hao},
  title = {PIE-NET: Parametric Inference of Point Cloud Edges},
  booktitle = NIPS,
  year = {2020}
}

@inproceedings{Liu2020DEF,
  author = {Matveev, Albert and Rakhimov, Ruslan and Artemov, Alexey and Bobrovskikh, Gleb and Egiazarian, Vage and Bogomolov, Emil and Panozzo, Daniele and Zorin, Denis and Burnaev, Evgeny},
  title = {DEF: Deep Estimation of Sharp Geometric Features in 3D Shapes},
  booktitle = SIGGRAPH,
  year = {2022}
}

@inproceedings{Zhang2021EDCNet,
  author = {Bazazian, Dena and Par{\'e}s, M. Eul{\`a}lia},
  title = {EDC-Net: Edge Detection Capsule Network for 3D Point Clouds},
  booktitle = APS,
  year = {2021}
}

@inproceedings{Guo2024NerVE,
  author = {Zhu, Xiangyu and Du, Dong and Chen, Weikai and Zhao, Zhiyou and Nie, Yinyu and Han, Xiaoguang},
  title = {NerVE: Neural Volumetric Edges for Parametric Curve Extraction from Point Clouds},
  booktitle = CVPR,
  year = {2023}
}

@inproceedings{Sharma2018CSGNet,
  author = {Sharma, Gopal and Goyal, Rishabh and Liu, Difan and Kalogerakis, Evangelos and Maji, Subhransu},
  title = {CSGNet: Neural Shape Parser for Constructive Solid Geometry},
  booktitle = CVPR,
  year = {2018}
}

@inproceedings{Chen2020CAPRINet,
  author = {Chen, Y. and et al.},
  title = {CAPRI-Net: Learning Compact CAD Shapes with Adaptive Primitive Assembly},
  booktitle = CVPR,
  year = {2022}
}

@inproceedings{Ji2020UCSGNet,
  author    = {Kania, Kacper and Zięba, Maciej and Kajdanowicz, Tomasz},
  title = {UCSG-Net: Unsupervised Discovering of Constructive Solid Geometry Tree},
  booktitle = NIPS,
  year = {2020}
}

@inproceedings{Li2020CSGStump,
  author    = {Ren, Daxuan and Zheng, Jianmin and Cai, Jianfei and Li, Jiatong and Jiang, Haiyong and Cai, Zhongang and Zhang, Junzhe and Pan, Liang and Zhang, Mingyuan and Zhao, Haiyu and Yi, Shuai},
  title = {CSG-Stump: A Learning Friendly CSG-Like Representation for Interpretable Shape Parsing},
  booktitle = ICCV,
  year = {2021}
}

@article{Gao2023DualCSG,
  author    = {Yu, Fenggen and Chen, Qimin and Tanveer, Maham and Mahdavi-Amiri, Ali and Zhang, Hao},
  title = {DualCSG: Learning Dual CSG Trees for General and Compact CAD Modeling},
  journal = {ArXiv},
  year = {2023},
  volume={abs/2301.11497},
  url={https://api.semanticscholar.org/CorpusID:256358481}
}

@inproceedings{Huang2024SECADNet,
  author = {Li, Pu and Guo, Jianwei and Zhang, Xiaopeng and Yan, Dong-Ming},
  title = {SECAD-Net: Self-Supervised CAD Reconstruction by Learning Sketch-Extrude Operations},
  booktitle = CVPR,
  year = {2023}
}

@inproceedings{Chen2025SfmCAD,
  author = {Li, Pu and Guo, Jianwei and Li, Huibin and Benes, Bedrich and Yan, Dong-Ming},
  title = {SfmCAD: Unsupervised CAD Reconstruction by Learning Sketch-based Feature Modeling Operations},
  booktitle = CVPR,
  year = {2024}
}

@inproceedings{Willis2021BRepNet,
  author = {Lambourne, Joseph G. and Willis, Karl D. D. and Jayaraman, Pradeep Kumar and Sanghi, Aditya and Meltzer, Peter and Shayani, Hooman},
  title = {BRepNet: A Topological Message Passing System for Solid Models},
  booktitle = CVPR,
  year = {2021}
}

@inproceedings{Zhang2023ComplexGen,
  author = {Guo, Haoxiang and Liu, Shilin and Pan, Hao and Liu, Yang and Tong, Xin and Guo, Baining},
  title = {ComplexGen: CAD Reconstruction by B-Rep Chain Complex Generation},
  booktitle = SIGGRAPH,
  year = {2022}
}

@inproceedings{Wu2024SEDNet,
  author = {Li, Yuanqi and Liu, Shun and Yang, Xinran and Guo, Jianwei and Guo, Jie and Guo, Yanwen},
  title = {SED-Net: Surface and Edge Detection for Primitive Fitting of Point Clouds},
  booktitle = SIGGRAPH,
  year = {2023}
}

@article{qian2022pcernet,
  title={Deep Point Cloud Edge Reconstruction via Surface Patch Segmentation},
  author={Li, Yuanqi and Wang, Hongshen and Liu, Yansong and Huang, Jingcheng and Liu, Shun and Huang, Chenyu},
  journal={IEEE Transactions on Visualization and Computer Graphics},
  year={2025},
}

@inproceedings{Zhu2024SplitFit,
  author = {Liu, Yilin and Chen, Jiale and Pan, Shanshan and Cohen-Or, Daniel and Zhang, Hao and Huang, Hui},
  title = {Split-and-Fit: Learning B-Reps via Structure-Aware Voronoi Partitioning},
  booktitle = SIGGRAPH,
  year = {2024}
}

@inproceedings{Zhou2024Point2CAD,
  author = {Liu, Yujia and Obukhov, Anton and Wegner, Jan Dirk and Schindler, Konrad},
  title = {Point2CAD: Reverse Engineering CAD Models from 3D Point Clouds},
  booktitle = CVPR,
  year = {2024}
}

@inproceedings{Penner2017Soft3D,
  author = {Penner, Eric and Zhang, Li},
  title = {Soft 3D Reconstruction for View Synthesis},
  booktitle = TOG,
  year = {2017}
}

@inproceedings{Mildenhall2020NeRF,
  author = {Mildenhall, Ben and Srinivasan, Pratul P. and Tancik, Matthew and Barron, Jonathan T. and Ramamoorthi, Ravi and Ng, Ren},
  title = {NeRF: Representing Scenes as Neural Radiance Fields for View Synthesis},
  booktitle = ECCV,
  year = {2020}
}

@inproceedings{Barron2021MipNeRF,
  author = {Barron, Jonathan T. and Mildenhall, Ben and Tancik, Matthew and Hedman, Peter and Martin-Brualla, Ricardo and Srinivasan, Pratul P.},
  title = {Mip-NeRF: A Multiscale Representation for Anti-Aliasing Neural Radiance Fields},
  booktitle = ICCV,
  year = {2021}
}

@inproceedings{Wang2021NeuS,
  author = {Wang, Peng and Liu, Lingjie and Liu, Yuan and Theobalt, Christian and Komura, Taku and Wang, Wenping},
  title = {NeuS: Learning Neural Implicit Surfaces by Volume Rendering for Multi-view Reconstruction},
  booktitle = NIPS,
  year = {2021}
}

@inproceedings{Muller2022InstantNGP,
  author = {M{\"u}ller, Thomas and Evans, Alex and Schied, Christoph and Keller, Alexander},
  title = {Instant Neural Graphics Primitives with a Multiresolution Hash Encoding},
  booktitle = SIGGRAPH,
  journal = TOG,
  volume = {41},
  number = {4},
  pages = {102:1--102:15},
  year = {2022}
}

@inproceedings{Kerbl2023Gaussian,
  author = {Kerbl, Bernhard and Kopanas, Georgios and Leimk{\"u}hler, Thomas and Drettakis, George},
  title = {3D Gaussian Splatting for Real-Time Radiance Field Rendering},
  booktitle = SIGGRAPH,
  journal = TOG,
  volume = {42},
  number = {4},
  year = {2023}
}

@inproceedings{Wang2024SuGaR,
  author = {Gu{\'e}don, Antoine and Lepetit, Vincent},
  title = {SuGaR: Surface-Aligned Gaussian Splatting for Efficient 3D Mesh Reconstruction and High-Quality Mesh Rendering},
  booktitle = CVPR,
  year = {2024}
}

@inproceedings{Wetzstein2024TwoDGS,
  author = {Huang, Binbin and Yu, Zehao and Chen, Anpei and Geiger, Andreas and Gao, Shenghua},
  title = {2D Gaussian Splatting for Geometrically Accurate Radiance Fields},
  booktitle = SIGGRAPH,
  year = {2024}
}

@inproceedings{Wang2024SA3DGS,
  author = {Cen, Jiazhong and Fang, Jiemin and Yang, Chen and Xie, Lingxi and Zhang, Xiaopeng and Shen, Wei and Tian, Qi},
  title = {Segment Any 3D Gaussians},
  booktitle = AAAI,
  year = {2025}
}

@article{Xu2024SemanticGS,
  author = {Guo, Jun and Ma, Xiaojian and Fan, Yue and Liu, Huaping and Li, Qing},
  title = {Semantic Gaussians: Open-Vocabulary Scene Understanding with 3D Gaussian Splatting},
  journal = {ArXiv},
  year = {2024},
  volume={abs/2403.15624},
  url={https://api.semanticscholar.org/CorpusID:268680548}
}

@inproceedings{Li2024Feature3DGS,
  author = {Zhou, Shijie and Chang, Haoran and Jiang, Sicheng and Fan, Zhiwen and Zhu, Zehao and Xu, Dejia and Chari, Pradyumna and You, Suya and Wang, Zhangyang and Kadambi, Achuta},
  title = {Feature 3DGS: Supercharging 3D Gaussian Splatting to Enable Distilled Feature Fields},
  booktitle = CVPR,
  year = {2023}
}

@inproceedings{Zhang2025CurveAwareGS,
  author    = {Gao, Zhirui and Yi, Renjiao and Dai, Yaqiao and Zhu, Xuening and Chen, Wei and Zhu, Chenyang and Xu, Kai},
  title     = {Curve-Aware Gaussian Splatting for 3D Parametric Curve Reconstruction},
  booktitle = ICCV,
  year      = {2025}
}

@inproceedings{Wang2024SGCR,
  author    = {Yang, Xinran and Ji, Donghao and Li, Yuanqi and Guo, Jie and Guo, Yanwen and Xie, Junyuan},
  title     = {SGCR: Spherical Gaussians for Efficient 3D Curve Reconstruction},
  booktitle = CVPR,
  year      = {2025}
}

@inproceedings{Chen2024SketchSplat,
  author    = {Ying, Haiyang and Zwicker, Matthias},
  title     = {SketchSplat: 3D Edge Reconstruction via Differentiable Multi-View Sketch Splatting},
  booktitle = ICCV,
  year      = {2025}
}

@inproceedings{Liu2024HybridPartGS,
  author    = {Gao, Zhirui and Yi, Renjiao and Huang, Yuhang and Chen, Wei and Zhu, Chenyang and Xu, Kai},
  title     = {Self-Supervised Learning of Hybrid Part-aware 3D Representations of 2D Gaussians and Superquadrics},
  booktitle = ICCV,
  year      = {2025}
}

@InProceedings{Koch_2019_CVPR,
  author    = {Koch, Sebastian and Matveev, Albert and Jiang, Zhongshi and Williams, Francis and Artemov, Alexey and Burnaev, Evgeny and Alexa, Marc and Zorin, Denis and Panozzo, Daniele},
  title     = {ABC: A Big CAD Model Dataset for Geometric Deep Learning},
  booktitle = CVPR,
  year      = {2019}
}

@inproceedings{Ye_2023_CVPR,
  author    = {Ye, Yunfan and Yi, Renjiao and Gao, Zhirui and Zhu, Chenyang and Cai, Zhiping and Xu, Kai},
  title     = {NEF: Neural Edge Fields for 3D Parametric Curve Reconstruction From Multi-View Images},
  booktitle = CVPR,
  year      = {2023},
}

@InProceedings{Kirillov2023_SAM,
  author    = {Kirillov, Alexander and Mintun, Eric and Ravi, Nikhila and Mao, Hanzi and Rolland, Chloe and Gustafson, Laura and Xiao, Tete and Whitehead, Spencer and Berg, Alexander C. and Lo, Wan-Yen and Dollar, Piotr and Girshick, Ross},
  title     = {Segment Anything},
  booktitle = ICCV,
  year = {2023}
}

@inproceedings{collins2022abo,
  title     = {ABO: Dataset and Benchmarks for Real-World 3D Object Understanding},
  author    = {Collins, Jasmine and Goel, Shubham and Deng, Kenan and Luthra, Achleshwar and Xu, Leon and Gundogdu, Erhan and Zhang, Xi and Vicente, Tomas F. Yago and Dideriksen, Thomas and Arora, Himanshu and Guillaumin, Matthieu and Malik, Jitendra},
  booktitle = CVPR,
  year      = {2022}
}

@article{Fischler1981RANSAC,
  title     = {Random Sample Consensus: A Paradigm for Model Fitting with Applications to Image Analysis and Automated Cartography},
  author    = {Martin A. Fischler and Robert C. Bolles},
  journal   = {Communications of the ACM},
  year      = {1981},
}

@inproceedings{Liu2025HoLa,
author = {Liu, Yilin and others},  title = {HoLa: B-Rep Generation using a Holistic Latent Representation},
  booktitle = {SIGGRAPH},
  year = {2025}
}

@article{Alam2025GenCAD,
  author = {Alam, Md Ferdous and others},
  title = {GenCAD: Image-Conditioned Computer-Aided Design Generation with Transformer-Based Contrastive Representation and Diffusion Priors},
  journal = TMLR,
  year = {2025}
}

@inproceedings{Li2025CADDreamer,
  author = {Li, Yuan and Lin, Cheng and Liu, Yuan and Long, Xiaoxiao and Zhang, Chenxu and Wang, Ningna and Li, Xin and Wang, Wenping and Guo, Xiaohu},
  title = {CADDreamer: CAD Object Generation from Single-view Images},
  booktitle = CVPR,
  year = {2025}
}

@inproceedings{You2025Img2CAD,
  author = {You, Yang and Uy, Mikaela Angelina and Han, Jiaqi and Thomas, Rahul and Zhang, Haotong and Du, Yi and Chen, Hansheng and Engelmann, Francis and You, Suya and Guibas, Leonidas},
  title = {Img2CAD: Reverse Engineering 3D CAD Models from Images through VLM-Assisted Conditional Factorization},
  booktitle = {SIGGRAPH Asia},
  year = {2025}
}
}


\end{document}